\documentclass[11pt,a4paper]{article}

\usepackage[utf8]{inputenc}
\usepackage[T1]{fontenc}
\usepackage{amsmath,amssymb}
\usepackage{graphicx}
\usepackage{booktabs}
\usepackage{hyperref}
\usepackage[margin=1in]{geometry}
\usepackage{caption}
\usepackage{subcaption}
\usepackage{xcolor}
\usepackage[numbers]{natbib}

\hypersetup{
    colorlinks=true,
    linkcolor=blue,
    citecolor=blue,
    urlcolor=blue
}

\title{Reliability Under Randomness: An Empirical Analysis of Sparse and Dense Language Model Behavior Across Decoding Temperatures}
\author{Kabir Grover}
\date{December 2025}

\begin{document}

\maketitle

\begin{abstract}
The increasing prevalence of sparse Mixture-of-Experts (MoE) architectures in large language models raises important questions regarding their reliability under stochastic decoding. While conditional computation enables substantial gains in computational efficiency, it remains unclear whether the interaction between sparse routing and temperature-based sampling compromises output stability relative to dense architectures. This work investigates whether conditional computation in MoE models amplifies decoding-induced randomness, leading to reduced reliability as temperature increases.
We evaluate three representative models: OLMoE-7B (sparse base), Mixtral-8x7B (sparse instruction-tuned), and Qwen2.5-3B (dense instruction-tuned) on deterministic arithmetic reasoning tasks with objectively verifiable answers. Experiments span four decoding configurations, ranging from greedy decoding to T=1.0. Our evaluation encompasses accuracy, format compliance, output consistency across repeated generations, and confidence metrics, totaling 9,360 model generations.
Results demonstrate that the sparse instruction-tuned model exhibits stability comparable to the dense instruction-tuned model across all decoding temperatures, while the sparse base model shows systematic degradation as temperature increases. These findings indicate that instruction tuning, rather than architectural sparsity, is the primary determinant of robustness to decoding randomness on deterministic tasks. We discuss the implications of these results for deploying sparse language models in reliability-critical applications, highlighting scenarios in which sparse architectures can be safely adopted without sacrificing output stability.
\end{abstract}

\newpage
\tableofcontents
\newpage

\section{Introduction}

The deployment of language models in production environments necessitates careful consideration of the trade-off between deterministic and stochastic decoding strategies. Greedy decoding ensures reproducibility, whereas temperature sampling introduces diversity that may benefit certain applications. However, for tasks requiring precise outputs, including mathematical reasoning, structured generation, and factual retrieval, sampling-induced randomness may compromise reliability.

This consideration assumes particular significance with the emergence of \textbf{sparse Mixture-of-Experts (MoE) models}. Unlike dense transformers, which activate all parameters for every token, MoE architectures route each token to a subset of specialized ``expert'' sub-networks. For instance, Mixtral-8x7B comprises 46.7 billion total parameters while activating only approximately 12.9 billion per token, achieving competitive performance at substantially reduced computational cost.

The sparse routing mechanism introduces a fundamental question: \textit{does conditional computation in MoE models interact adversely with temperature sampling?} If routing decisions exhibit sensitivity to minor perturbations, and temperature introduces stochasticity in token selection, these effects might compound to render sparse models less reliable than their dense counterparts.

\subsection{Research Objective}

This work investigates the following hypothesis:

\begin{quote}
\textbf{Do sparse MoE models exhibit accelerated performance degradation or qualitatively different behavior compared to dense models as decoding temperature increases?}
\end{quote}

To address this question, we designed controlled experiments utilizing deterministic arithmetic tasks, specifically problems with unambiguous correct answers that enable precise measurement of accuracy, format compliance, and consistency across multiple experimental runs.

\subsection{Principal Findings}

The central finding of this investigation can be summarized as follows: \textbf{temperature sensitivity is primarily determined by instruction tuning rather than architectural sparsity}.

The sparse instruction-tuned model (Mixtral) demonstrated equivalent stability to the dense instruction-tuned model (Qwen), with both maintaining consistent accuracy from greedy decoding through $T=1.0$. The only model exhibiting temperature-dependent degradation was the sparse \textit{base} model (OLMoE), which lacked instruction tuning.

These results suggest that instruction tuning confers robustness to temperature variation independent of architecture. The routing variance inherent to MoE models does not appear to compound with sampling randomness for well-tuned models on deterministic tasks.

The remainder of this paper is organized as follows: Section~\ref{sec:background} provides technical background on language model decoding, temperature scaling, and MoE architectures. Section~\ref{sec:setup} details the experimental methodology. Section~\ref{sec:results} presents empirical results. Section~\ref{sec:discussion} discusses implications and limitations. Section~\ref{sec:conclusion} concludes with recommendations for practitioners.

\section{Background}
\label{sec:background}

\subsection{Autoregressive Language Model Generation}

Autoregressive language models generate text sequentially, one token at a time. Given a sequence of tokens $x_1, x_2, \ldots, x_t$, the model predicts a probability distribution over the vocabulary for the subsequent token:

\begin{equation}
P(x_{t+1} \mid x_1, \ldots, x_t)
\end{equation}

This distribution is obtained by applying the softmax function to the model's output logits:

\begin{equation}
P(x_{t+1} = v) = \frac{\exp(z_v)}{\sum_{v'} \exp(z_{v'})}
\end{equation}

where $z_v$ denotes the logit (pre-softmax score) for vocabulary item $v$.

\subsection{Temperature Scaling}

Temperature scaling modifies the output distribution prior to sampling. With temperature parameter $T$, the probability becomes:

\begin{equation}
P_T(x_{t+1} = v) = \frac{\exp(z_v / T)}{\sum_{v'} \exp(z_{v'} / T)}
\end{equation}

The temperature parameter produces the following effects:

\begin{itemize}
    \item \textbf{$T \rightarrow 0$ (Greedy decoding):} The distribution becomes increasingly peaked, with the highest-probability token receiving nearly all probability mass, resulting in deterministic selection.
    \item \textbf{$T = 1$:} The model's original learned distribution is utilized without modification.
    \item \textbf{$T > 1$:} The distribution flattens, increasing the probability of lower-ranked tokens being selected.
\end{itemize}

Holtzman et al.~\cite{holtzman2020} demonstrated that sampling from the full distribution (even at $T=1$) can produce degenerate text due to the unreliable tail, meaning tokens with very low probability that are occasionally selected. Methods such as top-$k$ and nucleus (top-$p$) sampling address this issue by truncating the distribution. Temperature remains the primary mechanism for controlling the diversity-determinism trade-off.

The present experiments evaluate four temperature configurations: greedy ($T=0$), $T=0.1$, $T=0.5$, and $T=1.0$, spanning fully deterministic decoding to the model's unmodified distribution.

\subsection{Mixture-of-Experts Architecture}

The Mixture-of-Experts (MoE) architecture~\cite{shazeer2017, fedus2022} replaces the dense feed-forward network (FFN) in transformer blocks with multiple parallel ``expert'' networks and a routing mechanism.

A standard transformer FFN computes:
\begin{equation}
\text{FFN}(x) = W_2 \cdot \text{ReLU}(W_1 \cdot x)
\end{equation}

An MoE layer with $N$ experts instead computes:
\begin{equation}
\text{MoE}(x) = \sum_{i=1}^{N} G(x)_i \cdot E_i(x)
\end{equation}

where $E_i$ represents the $i$-th expert network and $G(x)$ is a gating function that produces a sparse weight vector, with typically only the top-$k$ experts receiving non-zero weights.

The fundamental insight is that while the model has access to all experts during training, only a fraction are utilized during inference. This enables scaling model capacity (total parameters) without proportionally scaling computation (active parameters).

\textbf{Mixtral-8x7B} employs 8 experts per MoE layer with top-2 routing, wherein each token activates 2 of 8 experts. Total parameters: 47B. Active parameters per token: approximately 13B.

\subsection{Computational Efficiency: Dense versus Sparse Architectures}

From an inference computation perspective:

\begin{itemize}
    \item \textbf{Dense models:} Inference FLOPs $\approx$ 2 $\times$ Parameters per token
    \item \textbf{Sparse MoE models:} Inference FLOPs $\approx$ 2 $\times$ Active Parameters = 2 $\times$ Total Parameters $\times$ ($k$ / $N$)
\end{itemize}

where $k$ represents the number of active experts and $N$ represents the total number of experts.

This computational efficiency motivates substantial interest in MoE architectures. However, efficiency must be evaluated alongside reliability considerations. A potential concern arises: does sparse routing introduce variance that compounds with sampling randomness? If minor perturbations in hidden states result in different expert selections, the combination of temperature sampling and expert routing might produce less consistent outputs.

\subsection{Instruction Tuning}

Instruction tuning (or alignment) trains models to follow user instructions rather than merely predict the next token. Base models learn text completion; instruction-tuned models learn to execute specified tasks.

This distinction is relevant to the present experimental design, as models are instructed to ``output only the final number.'' A base model may ignore this instruction and provide explanatory reasoning. An instruction-tuned model should adhere to the specified format. Prior work has demonstrated that format compliance presents challenges even for capable models: Zhou et al.~\cite{zhou2023} found that GPT-4 achieves only approximately 80\% compliance on format-following benchmarks, with smaller models often achieving 30--40\%.

The distinction between base and instruction-tuned models proves central to our findings.

\section{Experimental Setup}
\label{sec:setup}

\subsection{Models}

We evaluate three transformer-based language models spanning sparse and dense architectures:

\begin{table}[htbp]
\centering
\begin{tabular}{@{}lllll@{}}
\toprule
\textbf{Model} & \textbf{Architecture} & \textbf{Total Params} & \textbf{Active Params} & \textbf{Type} \\
\midrule
OLMoE-7B & Sparse MoE & 6.9B & $\approx$ 1.3B & Base \\
Qwen2.5-3B & Dense & 3.1B & 3.1B & Instruct \\
Mixtral-8x7B & Sparse MoE (8 experts, top-2) & 47B & $\approx 13\,\text{B}$ & Instruct \\
\bottomrule
\end{tabular}
\caption{Models evaluated in this study.}
\label{tab:models}
\end{table}

\textbf{OLMoE-7B}~\cite{muennighoff2024} is a sparse Mixture-of-Experts model with 6.9 billion parameters and approximately 1 billion active per token. The \textit{base} version (without instruction tuning) is employed to examine sparse model behavior under varying temperatures in the absence of alignment.

\textbf{Qwen2.5-3B} is a dense transformer with 3 billion parameters, all active during inference. The instruction-tuned variant is utilized, representing a well-aligned dense model.

\textbf{Mixtral-8x7B}~\cite{jiang2024} is a large sparse MoE model with 8 experts per layer and top-2 routing. With 46.7 billion total parameters but only approximately 12.9 billion active per token, it achieves strong performance at reduced computational cost. The instruction-tuned variant is employed.

Due to memory constraints, OLMoE and Mixtral were executed with 4-bit quantization (bitsandbytes NF4). Qwen2.5-3B operates in full precision.

\textbf{Computational Resources:} Experiments were conducted on an NVIDIA A100 GPU with 40GB VRAM. Total inference time for 9,360 generations was approximately 3.5 hours.

\subsection{Tasks}

Two deterministic reasoning tasks with objectively verifiable answers are employed:

\textbf{Variable Binding (120 examples).} Track variable assignments through sequential operations:
\begin{verbatim}
  a=3, b=2, c=a+b, d=c+1. d = ?    (Answer: 6)
\end{verbatim}

\textbf{Multi-step Arithmetic (120 examples).} Evaluate expressions requiring order of operations:
\begin{verbatim}
  (45-12)*3+7 = ?    (Answer: 106)
\end{verbatim}

Both tasks possess unambiguous correct answers. Prompts instruct the model to output \textit{only} the final number, enabling measurement of both accuracy and format compliance.

\subsection{Decoding Configurations}

\begin{table}[htbp]
\centering
\begin{tabular}{@{}llll@{}}
\toprule
\textbf{Setting} & \textbf{Sampling} & \textbf{Temperature} & \textbf{Runs} \\
\midrule
Greedy & No & -- & 1 \\
T=0.1 & Yes & 0.1 & 4 \\
T=0.5 & Yes & 0.5 & 4 \\
T=1.0 & Yes & 1.0 & 4 \\
\bottomrule
\end{tabular}
\caption{Decoding configurations evaluated.}
\label{tab:settings}
\end{table}

Greedy decoding is deterministic, requiring only a single run. Sampling configurations utilize multiple runs to assess consistency. No top-$k$ or top-$p$ truncation is applied; temperature constitutes the sole source of randomness.

\textbf{Total experimental scale:} 3 models $\times$ 240 examples $\times$ 13 runs = 9,360 generations.

\subsection{Evaluation Metrics}

\textbf{Accuracy.} The first integer is extracted from each output and compared to the ground truth. Evaluation is lenient: ``The answer is 106'' is marked correct (though non-compliant).

\textbf{Compliance.} Binary assessment of whether the output consists \textit{exclusively} of a numeric value. ``106'' is compliant; ``106. Here is my reasoning...'' is non-compliant.

\textbf{Consistency.} Standard deviation of accuracy across runs at each temperature setting. Lower values indicate more consistent behavior.

\textbf{Confidence.} Mean log-probability of generated tokens. More negative values indicate lower model confidence.

\section{Results}
\label{sec:results}

\subsection{Overall Accuracy}

\begin{figure}[htbp]
\centering
\includegraphics[width=0.9\textwidth]{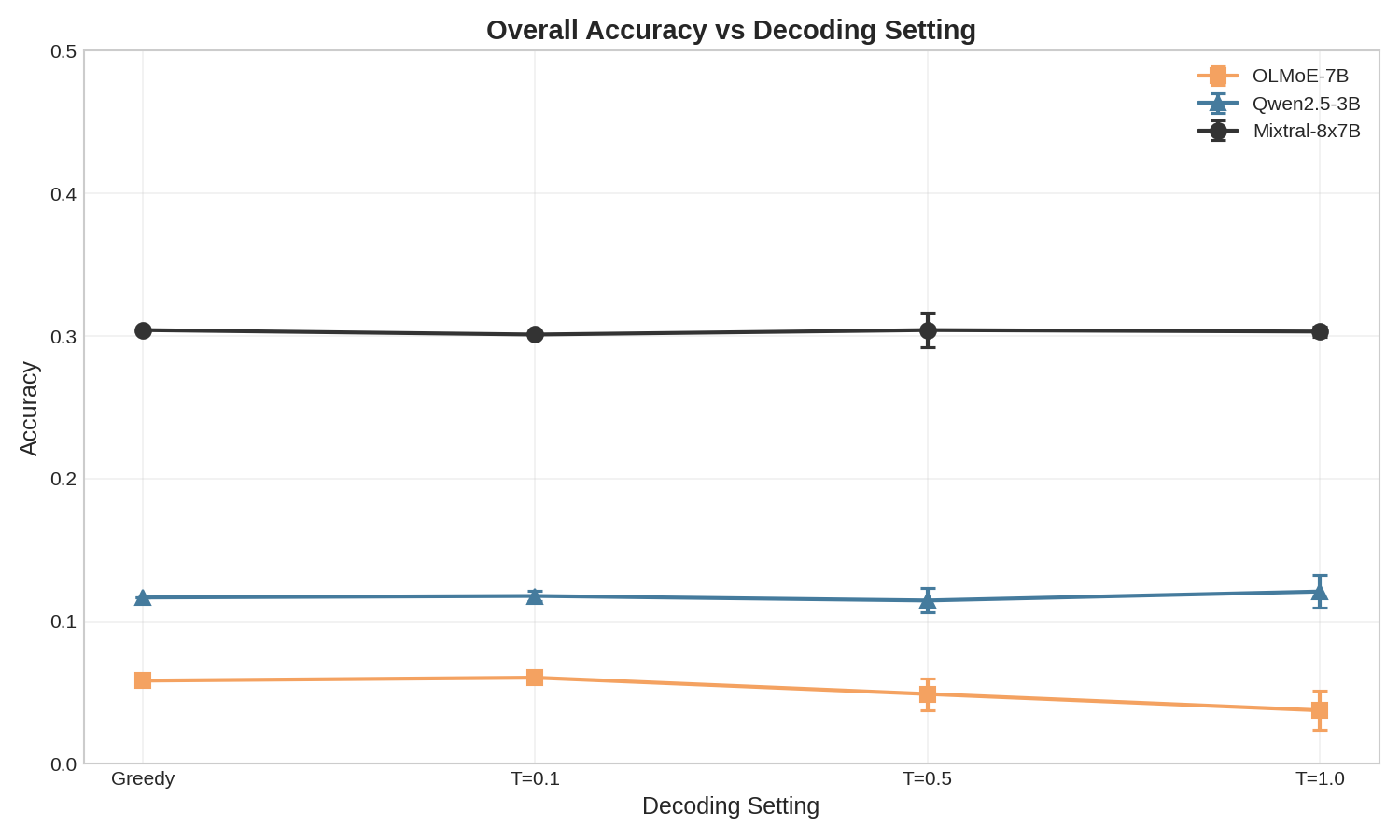}
\caption{Overall accuracy across decoding configurations. Error bars indicate standard deviation across runs.}
\label{fig:accuracy}
\end{figure}

Figure~\ref{fig:accuracy} presents accuracy as a function of temperature for all three models. The x-axis represents temperature settings from greedy (deterministic) to $T=1.0$ (maximum stochasticity). Error bars indicate variance across repeated runs.

\begin{table}[htbp]
\centering
\begin{tabular}{@{}lccccc@{}}
\toprule
\textbf{Model} & \textbf{Greedy} & \textbf{T=0.1} & \textbf{T=0.5} & \textbf{T=1.0} & \textbf{Trend} \\
\midrule
OLMoE-7B & 5.8\% & 6.0\% & 4.9\% & 3.8\% & $\downarrow$ Degradation \\
Qwen2.5-3B & 11.7\% & 11.8\% & 11.5\% & 12.1\% & $\rightarrow$ Stable \\
Mixtral-8x7B & 30.4\% & 30.1\% & 30.4\% & 30.3\% & $\rightarrow$ Stable \\
\bottomrule
\end{tabular}
\caption{Accuracy by model and temperature configuration.}
\label{tab:accuracy}
\end{table}

\textbf{OLMoE-7B} exhibits a pronounced downward trend, declining from 5.8\% accuracy with greedy decoding to 3.8

\textbf{Qwen2.5-3B} maintains stable performance, fluctuating minimally around 11--12\% accuracy across all configurations. Temperature exerts negligible effect on this model's performance.

\textbf{Mixtral-8x7B} similarly maintains stable performance, holding consistently at approximately 30\% accuracy. Despite being the largest model with the most complex routing mechanism, it exhibits no degradation.

Notably, Mixtral achieves the highest accuracy despite its sparse architecture. This observation provides initial evidence that sparsity alone does not compromise reliability; if architectural sparsity were problematic, Mixtral would be expected to exhibit the least stability rather than the highest accuracy.

\subsection{Architectural Comparison: Sparse versus Dense}

\begin{figure}[htbp]
\centering
\includegraphics[width=0.9\textwidth]{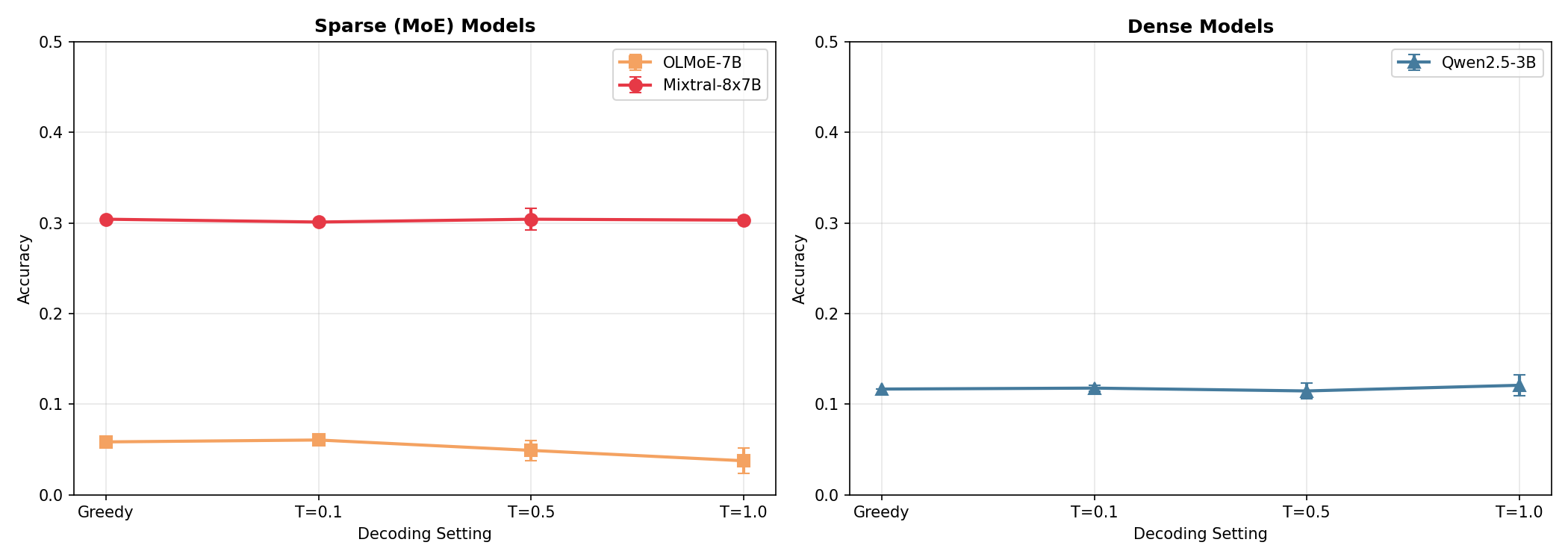}
\caption{Comparison of sparse models (left panel) versus dense models (right panel). OLMoE exhibits degradation while Mixtral maintains stability.}
\label{fig:sparse_dense}
\end{figure}

Figure~\ref{fig:sparse_dense} presents accuracy data partitioned by architecture type. The left panel displays the two sparse MoE models (OLMoE and Mixtral); the right panel displays the dense model (Qwen).

If sparse architecture induced temperature sensitivity, both models in the left panel would be expected to exhibit declining performance. The empirical results contradict this expectation:

\begin{itemize}
    \item \textbf{Left panel (Sparse models):} Mixtral remains stable across all temperatures; OLMoE exhibits degradation with increasing temperature.
    \item \textbf{Right panel (Dense model):} Qwen maintains stable performance across all temperatures.
\end{itemize}

The critical observation is that the two sparse models do not exhibit uniform behavior. If sparsity itself caused temperature sensitivity, both OLMoE and Mixtral would be expected to degrade. The distinguishing characteristic between OLMoE (which degrades) and the other models (which remain stable) is not architectural: OLMoE is a \textbf{base model} lacking instruction tuning, whereas Mixtral and Qwen are both \textbf{instruction-tuned}.

\subsection{Format Compliance}

\begin{figure}[htbp]
\centering
\includegraphics[width=0.9\textwidth]{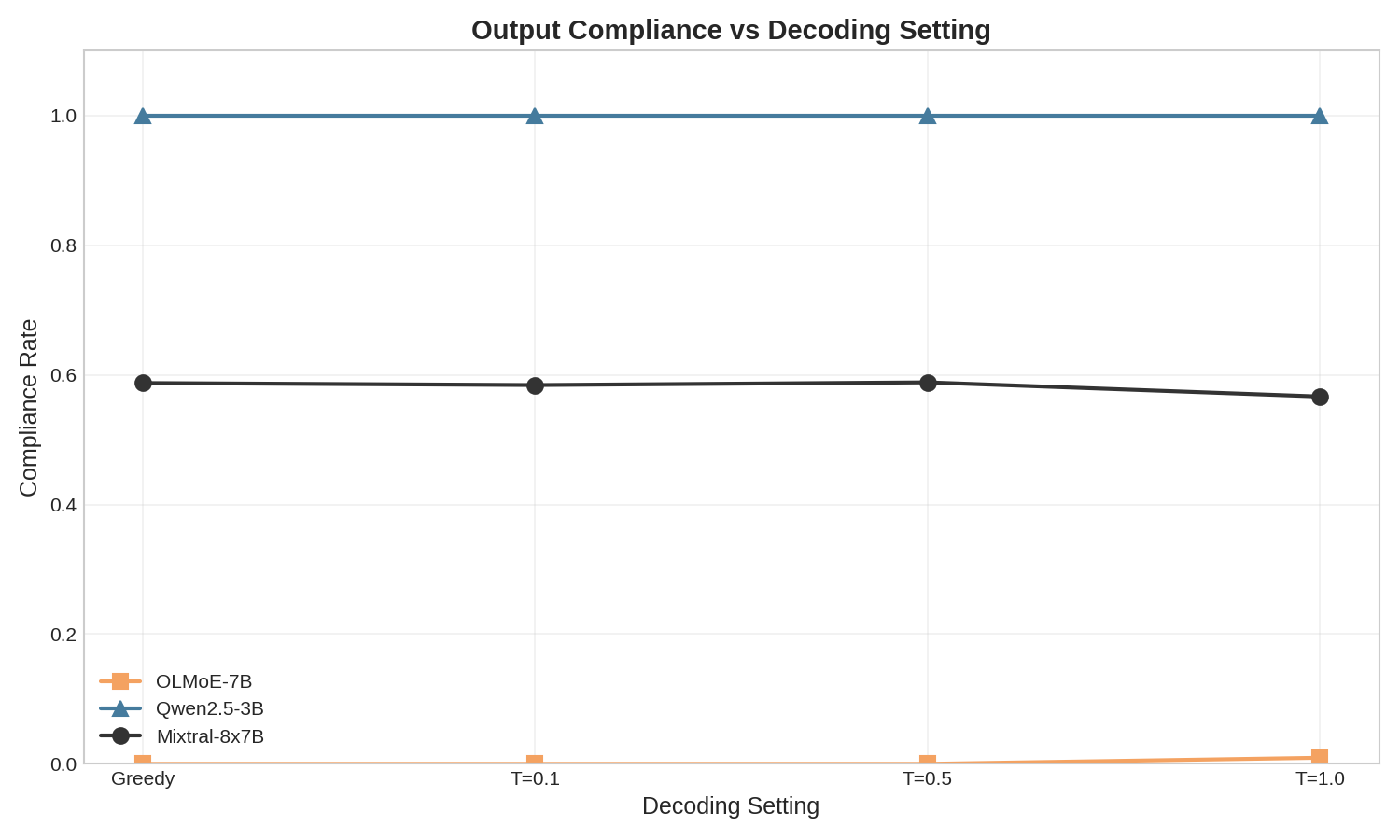}
\caption{Output format compliance across decoding configurations.}
\label{fig:compliance}
\end{figure}

\begin{table}[htbp]
\centering
\begin{tabular}{@{}lcccc@{}}
\toprule
\textbf{Model} & \textbf{Greedy} & \textbf{T=0.1} & \textbf{T=0.5} & \textbf{T=1.0} \\
\midrule
OLMoE-7B & 0\% & 0\% & 0\% & $\sim$1\% \\
Qwen2.5-3B & 100\% & 100\% & 100\% & 100\% \\
Mixtral-8x7B & 58.8\% & 58.4\% & 58.9\% & 56.7\% \\
\bottomrule
\end{tabular}
\caption{Compliance rates (outputting only a numeric value as instructed).}
\label{tab:compliance}
\end{table}

Format compliance results reveal three distinct behavioral patterns:

\textbf{OLMoE-7B} achieves \textbf{0\% compliance}. Across over 3,000 generations, this model never produced an output consisting solely of a numeric value. Every response included explanations, reasoning steps, or supplementary text. As a base model without instruction tuning, it fails to interpret and execute the format specification.

\textbf{Qwen2.5-3B} achieves \textbf{100\% compliance}. Every response consisted exclusively of a numeric value. When incorrect, the model still adhered to the specified format. This demonstrates complete understanding and execution of the instruction.

\textbf{Mixtral-8x7B} achieves \textbf{approximately 58\% compliance}. This instruction-tuned model typically adheres to the format but includes supplementary text in approximately 42\% of responses. Compliance decreases marginally at $T=1.0$ (56.7\%), suggesting higher stochasticity slightly increases the probability of format violations.

Format compliance provides explanatory power for the accuracy patterns observed. OLMoE's 0\% compliance results in verbose outputs with increased opportunity for errors. Qwen's 100\% compliance produces concise, focused outputs.

\subsection{Failure Mode Analysis}

\begin{figure}[htbp]
\centering
\includegraphics[width=0.95\textwidth]{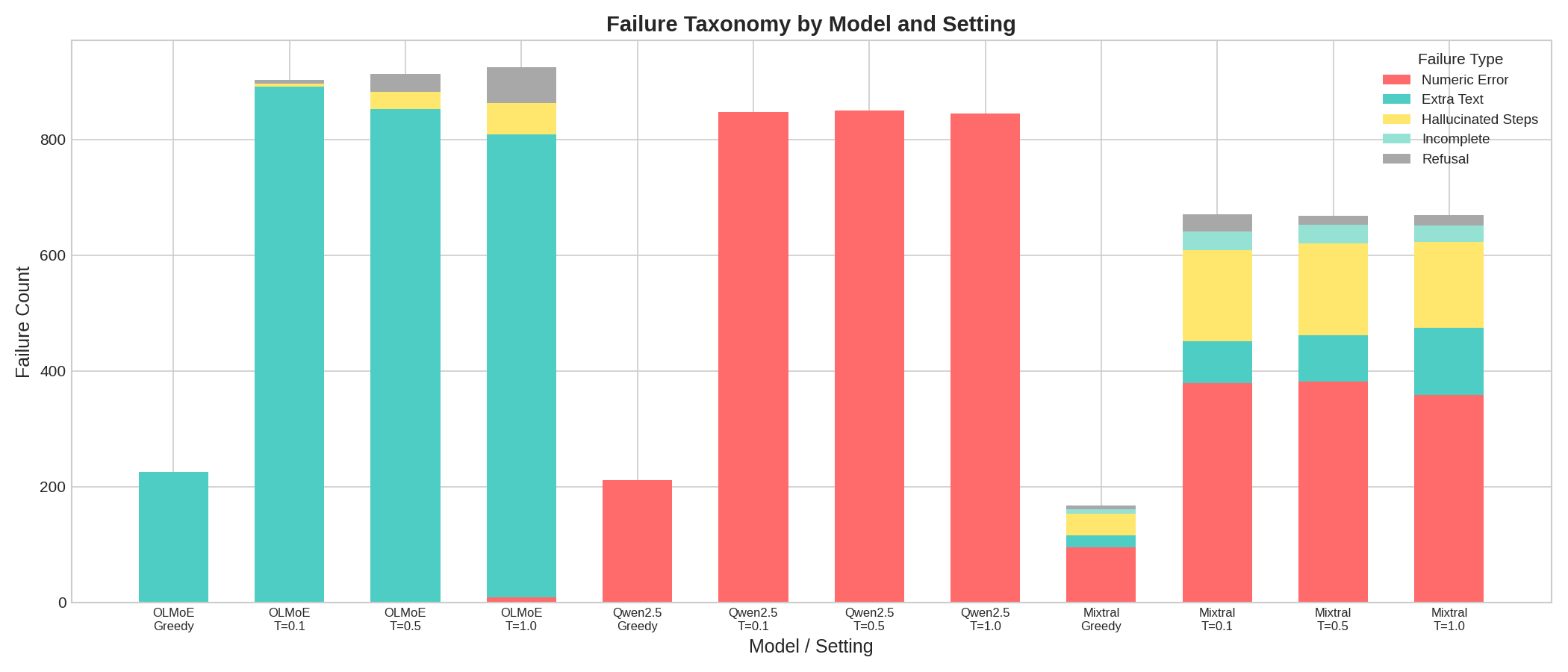}
\caption{Distribution of failure types by model and decoding configuration.}
\label{fig:failures}
\end{figure}

Figure~\ref{fig:failures} presents a taxonomy of failure modes, categorized as follows:
\begin{itemize}
    \item \textbf{Numeric error:} Correct format but incorrect numeric value
    \item \textbf{Extra text:} Supplementary explanations or text beyond the numeric answer
    \item \textbf{Hallucinated steps:} Fabricated reasoning or intermediate computations
    \item \textbf{Refusal:} Explicit refusal to provide an answer
    \item \textbf{Incomplete:} Truncated or empty output
\end{itemize}

\textbf{OLMoE-7B:} Failures are dominated by ``extra text'' (approximately 100\% at greedy). As temperature increases, small proportions of hallucinated steps and refusals emerge, but excessive verbosity remains the primary failure mode.

\textbf{Qwen2.5-3B:} 100\% of failures are classified as ``numeric error.'' The model consistently adheres to the format; failures occur exclusively due to computational errors. This represents predictable, well-characterized failure behavior.

\textbf{Mixtral-8x7B:} Mixed failure modes are observed: approximately 57\% numeric errors, 12--17\% extra text, and approximately 22\% hallucinated steps. Failure distribution remains relatively consistent across temperatures.

This analysis reveals a fundamental distinction in failure characteristics:

\begin{itemize}
    \item \textbf{Base model (OLMoE):} Failures arise from instruction non-compliance, a behavioral deficit.
    \item \textbf{Instruction-tuned models (Qwen, Mixtral):} Failures arise from computational errors, a capability limitation.
\end{itemize}

\subsection{Task-Specific Analysis: Variable Binding}

\begin{figure}[htbp]
\centering
\includegraphics[width=0.9\textwidth]{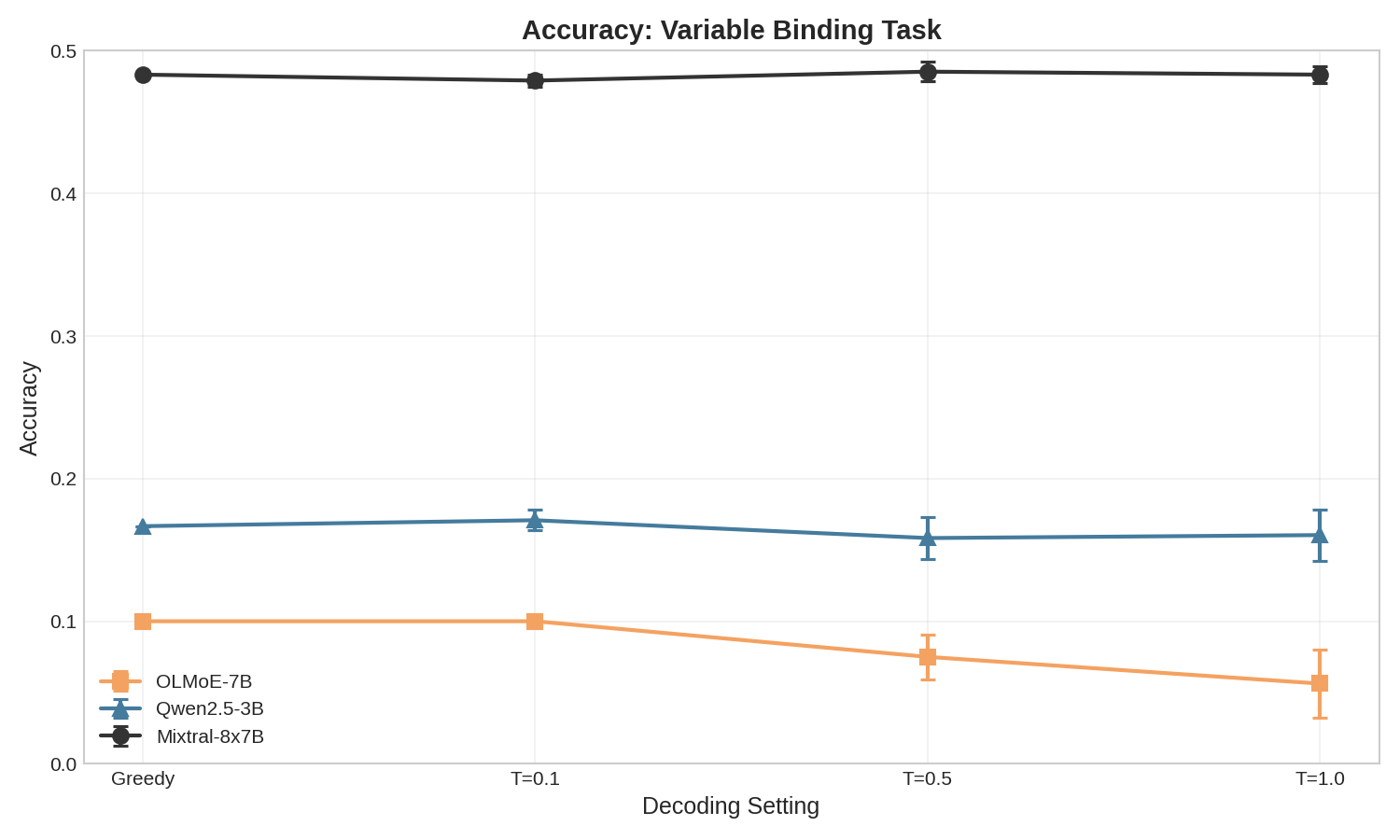}
\caption{Accuracy on the Variable Binding task across decoding configurations.}
\label{fig:accuracy_var_bind}
\end{figure}

Variable Binding requires tracking state across multiple sequential assignments, representing a distinct cognitive skill from pure arithmetic computation. Results maintain the pattern observed in aggregate analysis:

\begin{itemize}
    \item \textbf{Mixtral-8x7B} achieves the highest accuracy, demonstrating that sparse routing handles sequential variable tracking effectively.
    \item \textbf{Qwen2.5-3B} performs consistently across temperatures.
    \item \textbf{OLMoE-7B} continues to exhibit degradation with temperature, indicating that base model limitations persist across task types.
\end{itemize}

\subsection{Task-Specific Analysis: Multi-step Arithmetic}

\begin{figure}[htbp]
\centering
\includegraphics[width=0.9\textwidth]{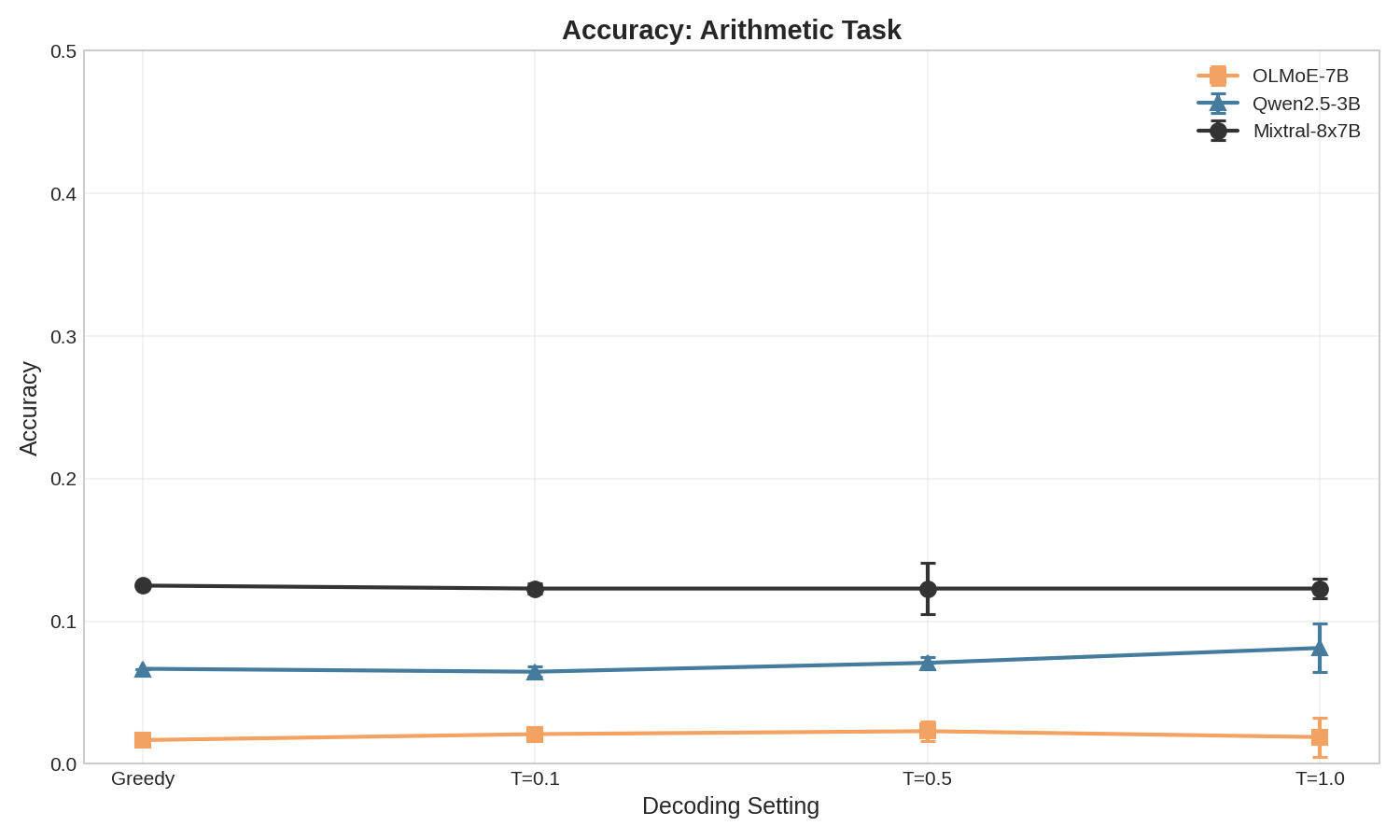}
\caption{Accuracy on the Multi-step Arithmetic task across decoding configurations.}
\label{fig:accuracy_arithmetic}
\end{figure}

Multi-step Arithmetic requires maintaining intermediate results and applying operations in correct precedence order. Results mirror the Variable Binding analysis:

\begin{itemize}
    \item The overall pattern replicates across tasks: instruction-tuned models remain stable while the base model degrades.
    \item \textbf{Mixtral-8x7B} again achieves highest accuracy, suggesting its larger active parameter count (13B) provides stronger arithmetic capabilities.
    \item \textbf{OLMoE-7B}'s degradation is consistent across both task types, confirming that temperature sensitivity is a general property of this base model rather than task-specific.
\end{itemize}

\subsection{Consistency Analysis}

\begin{figure}[htbp]
\centering
\includegraphics[width=0.9\textwidth]{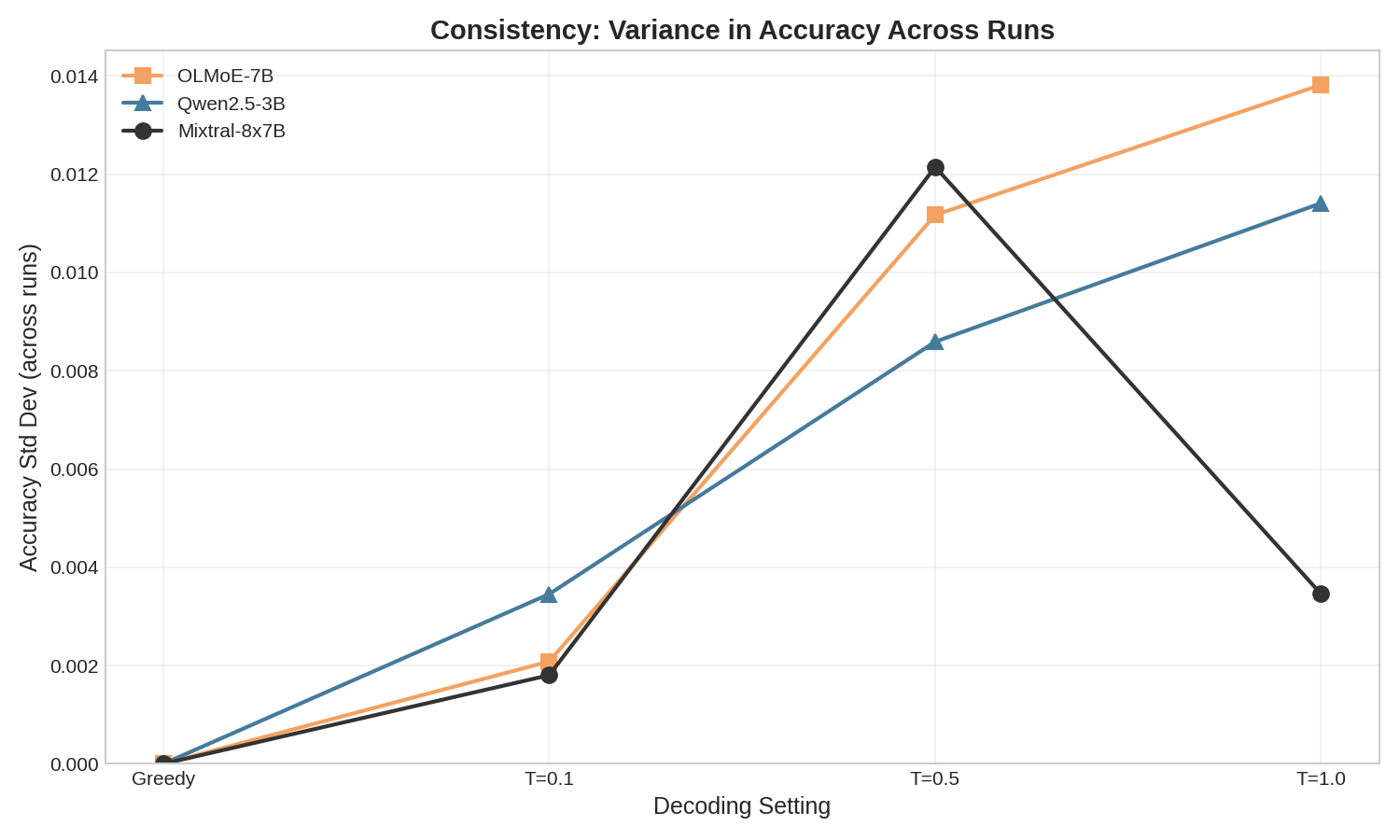}
\caption{Consistency (inverse of standard deviation) across multiple runs at each temperature configuration.}
\label{fig:consistency}
\end{figure}

Consistency quantifies the variance in accuracy across repeated runs at identical temperature settings. Higher values indicate more predictable behavior.

\begin{itemize}
    \item \textbf{Greedy decoding} ($T=0$) produces perfect consistency for all models, as expected given its deterministic nature.
    \item \textbf{Instruction-tuned models} (Mixtral and Qwen) maintain high consistency even at $T=1.0$, indicating training-induced robustness to sampling randomness.
    \item \textbf{OLMoE-7B} exhibits decreasing consistency with increasing temperature, indicating progressively less predictable outputs.
\end{itemize}

The combination of reduced accuracy \textit{and} reduced consistency at elevated temperatures renders base models particularly unsuitable for production deployment.

\subsection{Confidence Analysis}

\begin{figure}[htbp]
\centering
\includegraphics[width=0.9\textwidth]{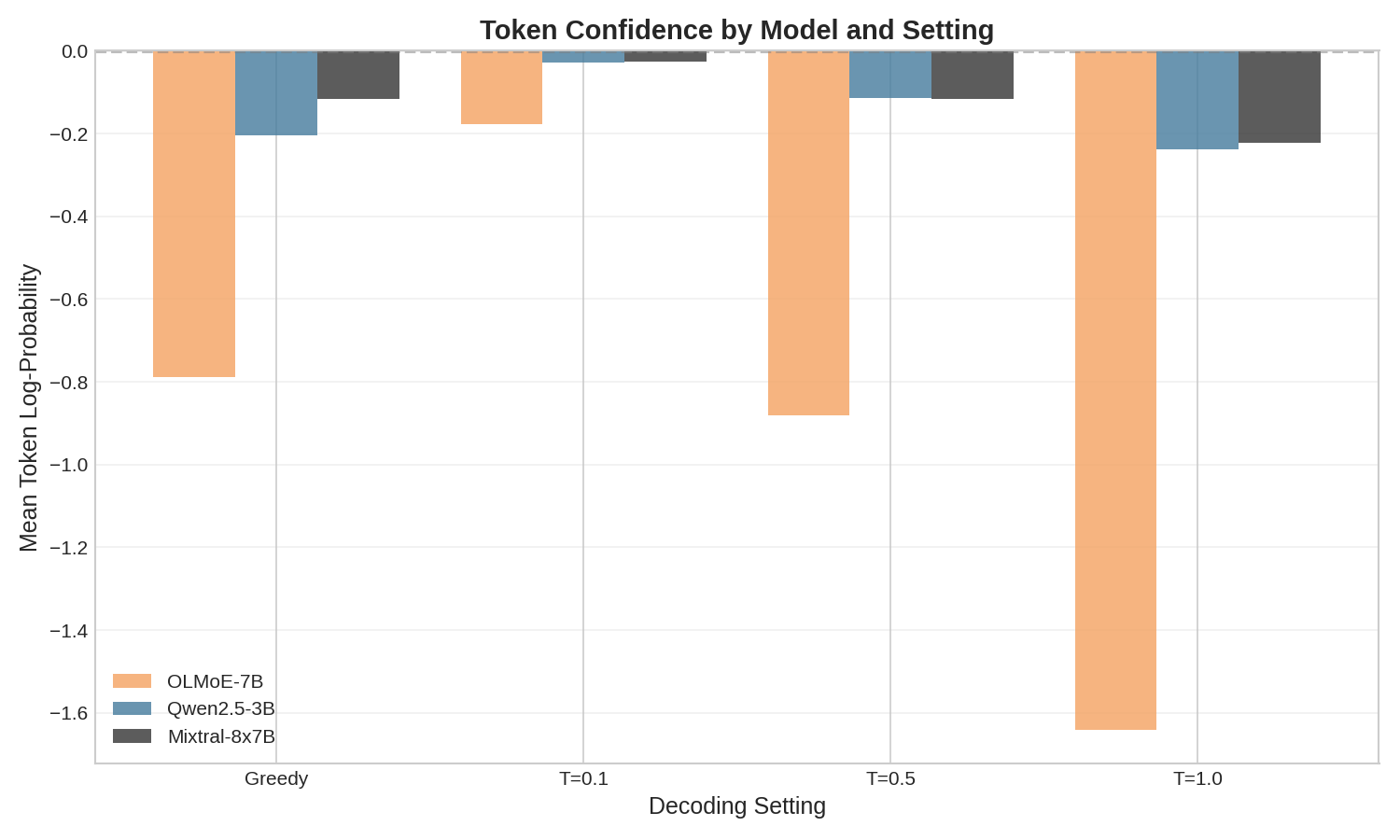}
\caption{Model confidence (mean log-probability of generated tokens) across decoding configurations.}
\label{fig:confidence}
\end{figure}

Confidence is quantified as the mean log-probability of generated tokens, with values approaching zero indicating higher confidence.

\begin{itemize}
    \item \textbf{Temperature naturally reduces confidence}, as elevated temperature flattens the probability distribution.
    \item \textbf{Mixtral-8x7B} exhibits the highest confidence overall, correlating with its highest accuracy.
    \item \textbf{Qwen2.5-3B} maintains moderate, stable confidence levels across temperatures.
    \item \textbf{OLMoE-7B} exhibits lower confidence, particularly at elevated temperatures.
\end{itemize}

Confidence correlates with instruction tuning quality: models with better format adherence exhibit greater output confidence. This suggests that well-aligned models possess more calibrated uncertainty estimates.

This observation has practical implications: token probability could serve as a reliability signal. For instruction-tuned models, high-confidence outputs (mean log-probability $> -0.5$) are more likely to be correct. Such confidence-based filtering would be less effective for base models, where confidence does not correlate as strongly with correctness.

\subsection{Results Summary}

\begin{itemize}
    \item \textbf{Accuracy:} OLMoE exhibited degradation with temperature (5.8\% $\rightarrow$ 3.8\%). Qwen and Mixtral maintained stability.
    \item \textbf{Task-specific performance:} Both Variable Binding and Arithmetic demonstrated consistent patterns: instruction-tuned models resist temperature degradation.
    \item \textbf{Architecture:} Sparse versus dense architecture did not determine stability. Mixtral (sparse) was the most stable model evaluated.
    \item \textbf{Compliance:} Qwen = 100\%, Mixtral $\approx$ 58\%, OLMoE = 0\%. Instruction tuning constitutes the differentiating factor.
    \item \textbf{Consistency:} Instruction-tuned models maintain consistent accuracy across runs; OLMoE exhibits increasing variance at elevated temperatures.
    \item \textbf{Confidence:} Higher confidence correlates with instruction tuning. Mixtral exhibits highest confidence; OLMoE exhibits lowest.
    \item \textbf{Failure modes:} Base models fail through format non-compliance. Instruction-tuned models fail through computational errors.
\end{itemize}

\section{Discussion}
\label{sec:discussion}

\subsection{Evaluation of Initial Hypothesis}

The initial hypothesis posited: \textbf{Do sparse MoE models degrade faster than dense models as decoding temperature increases?}

Based on the empirical results, the answer is: \textbf{temperature sensitivity is determined by instruction tuning rather than architecture}.

The observed patterns are as follows:
\begin{itemize}
    \item Mixtral-8x7B (sparse, instruction-tuned): stable across all temperatures
    \item Qwen2.5-3B (dense, instruction-tuned): stable across all temperatures
    \item OLMoE-7B (sparse, base): degradation with increasing temperature
\end{itemize}

The sparse instruction-tuned model (Mixtral) exhibited reliability equivalent to the dense instruction-tuned model (Qwen). The only model demonstrating degradation was the base model, which happened to utilize sparse architecture.

These results do not support the claim that ``sparse models degrade faster.'' Rather, among the models evaluated, instruction tuning appears to confer robustness to temperature variation independent of architectural choice.

\subsection{Experimental Confound}

A methodological limitation requires acknowledgment: the experimental design did not include a dense base model.

\begin{table}[htbp]
\centering
\begin{tabular}{@{}lcc@{}}
\toprule
 & \textbf{Base} & \textbf{Instruct} \\
\midrule
\textbf{Sparse} & OLMoE (degradation) & Mixtral (stable) \\
\textbf{Dense} & \textit{not evaluated} & Qwen (stable) \\
\bottomrule
\end{tabular}
\caption{Experimental coverage. The omitted cell limits causal inference.}
\label{tab:confound}
\end{table}

Without a dense base model, it is not possible to determine whether OLMoE's degradation results from:
\begin{enumerate}
    \item Sparse architecture (architectural effect), or
    \item Absence of instruction tuning (alignment effect), or
    \item Interaction of both factors
\end{enumerate}

Mixtral's stability (sparse, instruction-tuned) suggests sparsity alone is not problematic. However, a dense base model would be necessary to definitively exclude architecture as a contributing factor.

\subsection{Mechanistic Explanation for OLMoE Degradation}

The compliance analysis provides explanatory power for OLMoE's degradation pattern.

OLMoE achieves 0\% compliance; it never produces outputs consisting solely of numeric values. Every response includes explanatory content. The model does not interpret ``output only the final number'' as an instruction requiring execution.

As temperature increases:
\begin{itemize}
    \item Outputs become longer and more variable
    \item Increased verbosity provides more opportunity for errors
    \item Correct answers become obscured within explanatory text
    \item Occasional refusals or off-topic content emerge
\end{itemize}

Instruction-tuned models avoid this failure mode. Qwen produces exactly the requested format (100\% compliance). Mixtral predominantly does so (approximately 58\%). Concise, focused outputs limit the scope for temperature-induced drift.

\subsection{Probability Concentration Hypothesis}

The differential temperature sensitivity between base and instruction-tuned models may be explained through \textbf{probability concentration}.

When a model has high confidence in the correct response, it assigns extremely high probability to the correct token sequence (e.g., 99\%). Even at $T=1.0$, sampling will select the correct answer with high frequency.

For OLMoE, the issue is not probability concentration on incorrect answers; instead, the model does not understand the task format. Its probability mass is distributed across multiple plausible continuations (explanations, reasoning steps, tangential content). Temperature increases this diversity, producing less predictable outputs.

In contrast, Mixtral and Qwen have been trained to concentrate probability mass on the specified behavior: output a numeric value and terminate. This concentrated distribution exhibits robustness to sampling noise.

\subsection{Limitations}

\textbf{Missing comparison.} Absence of a dense base model prevents complete separation of architectural effects from instruction tuning effects. An ideal experimental design would include a model such as Llama2-7B-base to complete the 2$\times$2 factorial design.

\textbf{Model size variation.} Comparison spans 1B active parameters (OLMoE) versus 3B (Qwen) versus 13B active (Mixtral). Size differences may partially explain accuracy variations.

\textbf{Limited task scope.} Only arithmetic tasks were evaluated. Open-ended generation or creative tasks might exhibit different patterns.

\textbf{Single base model.} OLMoE represents one base model. Other base models might exhibit different behavior.

\textbf{Quantization effects.} OLMoE and Mixtral utilized 4-bit quantization, which could affect results. While quantization may influence absolute accuracy, relative temperature trends within models should be preserved.

\subsection{Robust Findings}

Despite methodological limitations, several findings are well-supported:

\begin{enumerate}
    \item \textbf{Sparse instruction-tuned models can achieve stability comparable to dense instruction-tuned models.} Mixtral exhibited no degradation with temperature; it was the most stable and accurate model evaluated.

    \item \textbf{Instruction tuning substantially affects compliance.} The gap between 0\% (OLMoE) and 100\% (Qwen) compliance demonstrates the magnitude of alignment's effect on format adherence.

    \item \textbf{Failure modes differ systematically by model type.} Base models fail through instruction non-compliance (excessive text). Instruction-tuned models fail through computational errors (incorrect numeric values).
\end{enumerate}

For practitioners: when deploying well-tuned models, moderate temperature (up to $T=1.0$) is unlikely to substantially compromise reliability on deterministic tasks. Model training quality is more consequential than architectural choice.

\section{Conclusion}
\label{sec:conclusion}

This investigation addressed the question: do sparse MoE models exhibit accelerated degradation compared to dense models as decoding temperature increases?

Following 9,360 generations across three models and four temperature configurations, the principal finding is: \textbf{instruction tuning, rather than architectural sparsity, determines temperature robustness}.

\subsection{Principal Conclusions}

\begin{enumerate}
    \item \textbf{Sparse architecture does not induce temperature sensitivity.} Mixtral (sparse, instruction-tuned) maintained stable performance across all temperature settings, matching or exceeding the dense model in reliability.

    \item \textbf{Instruction tuning confers robustness.} Both instruction-tuned models (sparse and dense) exhibited stability. The base model exhibited degradation. This suggests alignment training provides an anchor that maintains consistent behavior despite sampling stochasticity.

    \item \textbf{Compliance is binary.} OLMoE (base) achieved 0\% compliance; Qwen (instruction-tuned) achieved 100\%. This gap is attributable to training methodology rather than architecture.

    \item \textbf{Failure modes are systematic.} Base models fail through format non-compliance. Instruction-tuned models fail through computational errors. These represent distinct problems requiring different interventions.
\end{enumerate}

\subsection{Practical Recommendations}

For practitioners deploying models on deterministic tasks, moderate temperature (up to $T=1.0$) is unlikely to substantially compromise reliability when using well-tuned models. Model training quality is more consequential than sparse versus dense architecture.

However, for applications requiring strict format compliance (JSON output, structured responses), lower temperature or greedy decoding remains advisable. Even well-tuned models may exhibit increased format violations at elevated temperatures.

\subsection{Future Work}

An improved experimental design would evaluate all four combinations in the 2$\times$2 factorial:

\begin{itemize}
    \item Sparse + Base
    \item Sparse + Instruction-tuned
    \item Dense + Base
    \item Dense + Instruction-tuned
\end{itemize}

Additionally, controlling for model size through matched active parameters (e.g., 3B sparse versus 3B dense) would strengthen causal inference.

The present findings suggest that concerns regarding MoE model reliability under temperature sampling may be unwarranted, provided models are appropriately instruction-tuned.

\section*{Code Availability}

All code and evaluation data are publicly available at: \url{https://github.com/kabirrgrover/lm-reliability-sparsity}

\bibliographystyle{plainnat}
\bibliography{references}

\end{document}